\documentclass[letterpaper, 10 pt, conference]{ieeeconf}  

\IEEEoverridecommandlockouts                              

\usepackage{graphics} 
\usepackage{epsfig} 
\usepackage{stfloats} 
\usepackage{mathptmx} 
\usepackage{times} 
\usepackage{amsmath} 
\usepackage{amssymb}  

\usepackage{multirow} 
\usepackage{hyperref} 
\hypersetup{colorlinks=true} 

\usepackage[ruled,linesnumbered]{algorithm2e} 

\title{\LARGE \bf
Robotic Desk Organization: A Multi-Primitive Approach to Manipulating Heterogeneous Objects via Environmental Constraints
}

\author{Yi Dong$^{}$, 
Yangjun Liu$^{}$, 
Jinjun Duan$^{}$, 
Yang Li$^{}$, 
Zhendong Dai$^{}$ 
\thanks{This work has been submitted to the IEEE for possible publication. Copyright may be transferred without notice, after which this version may no longer be accessible.}
}

\begin{document}

\maketitle
\thispagestyle{empty}
\pagestyle{empty}

\begin{abstract}

Desktop organization remains challenging for service robots because of heterogeneous objects and diverse manipulation objectives, such as collection and stacking.
In this article, a task-oriented framework is presented for organizing planar rigid and deformable objects on desks.
A perception pipeline was developed that augments existing datasets with uncommon desktop items and makes geometry-based pose and keypoint estimation possible, along with the detection of environmental constraints, such as table edges.
To handle diverse manipulation requirements, environment-assisted primitives are used, including contact-based grasping for small objects, edge-based push–grasping for planar rigid objects, and levering-based grasping for planar deformable objects. These primitives leverage environmental and interobject constraints to improve robustness.
A task planner was designed to integrate these primitives into multiobject organization. 
Sufficient real-world experiments demonstrate the effectiveness and robustness of the proposed framework.
This research provides practical manipulation primitives for planar rigid and deformable objects, highlighting the role of environmental and interobject constraints in complex multiobject manipulation tasks.
\href{https://github.com/manipulation20/robotic-desk-organization}{Code} and \href{https://youtu.be/3n3nDDh0HTA}{video} are available online.

\end{abstract}

\section{INTRODUCTION}
\begin{figure}[htbp]
  \centering
  \includegraphics[width=8cm]{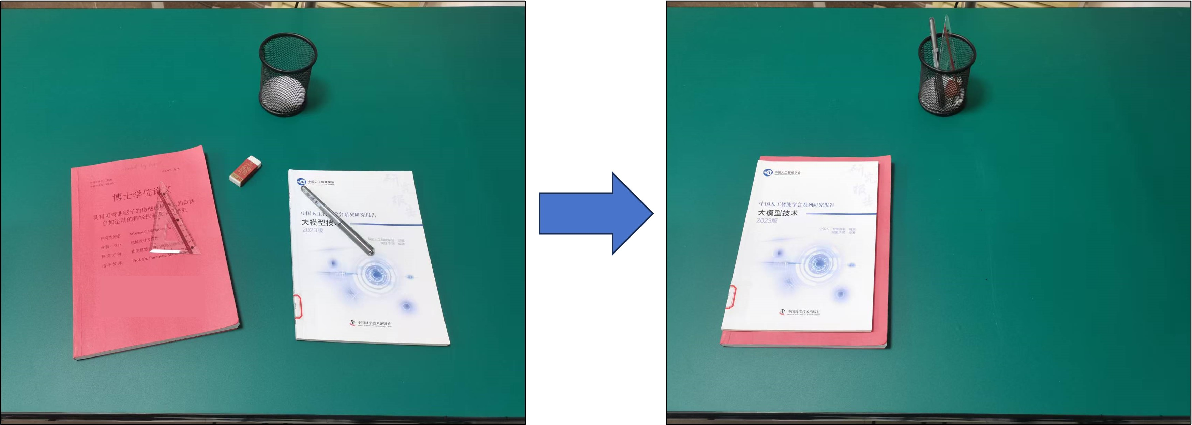}
  \caption{Initial and final states of the desktop organization task. }
  \label{Fig0}
\end{figure}

Tabletop organization is a fundamental capability for service robots and a representative task for embodied intelligence in real-world environments.
While prior studies have explored applications such as dining table setup \cite{zhai2024sg} and general object rearrangement \cite{hu2025planning} \cite{gao2022fast}, desk organization—particularly in office or study settings—remains underexplored.
A crucial challenge lies in the diversity of the objects involved.
Desk environments typically include small items (e.g., lead cases), planar rigid objects (e.g., rulers), and deformable objects (e.g., paper sheets and books) that exhibit significantly different physical properties and manipulation requirements (Fig. \ref{Fig0}).
Moreover, unlike conventional rearrangement tasks, desk organization often involves structured operations, such as stacking and category-based storage, which require task-oriented manipulation strategies.

In this study, for small objects, contact-based grasping methods were adopted that were proposed in prior work \cite{eppner2015exploitation}.
Beyond directly applying existing techniques, the robustness of such methods with respect to object height and grasping location, which are critical factors affecting grasp success in dense desktop settings, was investigated systematically.
For planar rigid objects, previous approaches relied on either specialized strategies (e.g., prying \cite{zhang2022prying} or scooping \cite{cha2024high} \cite{he2021scooping} \cite{babin2019stable}) or environmental constraints, such as table edges \cite{bimbo2019exploiting} \cite{hang2019pre}.
In this work, environment-assisted manipulation was extended by exploiting not only desk edges but also object–object constraints, such as book edges, to facilitate reliable grasping of rulers.

For deformable objects, thin materials, such as paper, are commonly grasped by inducing local deformation, similar to cloth manipulation \cite{borras2020grasping}.
However, unlike cloth, paper stiffness increases with thickness, which significantly affects graspability.
The influences of paper thickness and grasping location on the grasp success were analyzed explicitly.
For thicker objects, such as books, a prying-based strategy was adopted instead that treats them as quasi-rigid while leveraging their limited deformability.

Building on these object-specific primitives, a task planner tailored to desk organization was developed, making structured operations possible in realistic scenarios.
The main contributions of this study are as follows.
1) Previous stationery datasets were extended by incorporating uncommon objects, such as triangular rulers and lead cases.
2) Task-oriented manipulation primitives were designed for small, planar rigid, and deformable objects, and their effective workspaces were evaluated systematically.
3) A desk organization task planner was developed that integrates these primitives, and it was validated through extensive experiments.
4) It was demonstrated that, in addition to commonly used environmental constraints, interobject contacts (e.g., the edge of a book) in multiobject manipulation can also serve as effective resources.


\section{RELATED WORK}

\subsection{Robotic Table Organization}

Tabletop organization has been widely studied using prehensile \cite{gao2025orla} \cite{hu2025planning} \cite{zhai2024sg} \cite{cai2024visual} \cite{yuan2024robopoint} \cite{labbe2020monte}, nonprehensile \cite{dengler2022learning} \cite{vieira2022persistent} \cite{huang2021visual} \cite{song2020multi} \cite{song2019object}, or hybrid manipulation primitives \cite{huang2024toward}  \cite{tang2023selective}.
Most studies have focused on improving task and motion planning efficiency, such as reducing the execution steps and planning time \cite{liu2021ocrtoc} \cite{batra2020rearrangement}.
Nonprehensile actions, particularly pushing, are often introduced to assist in completing manipulation tasks, making object separation \cite{vieira2022persistent} \cite{huang2021visual}, reorientation \cite{huang2024toward}, and efficient rearrangement \cite{song2019object} \cite{song2020multi} possible.
However, these approaches primarily address general rearrangement tasks and rely heavily on pushing primitives, which are insufficient for handling hard-to-grasp objects or structured operations, such as stacking.
By contrast, this research focused on desk organization in realistic settings, requiring the coordinated use of diverse primitives for heterogeneous objects and emphasizing structured manipulation.

\subsection{Planner Objects Manipulation}

Grasping planar objects remains challenging for parallel grippers.
Established methods can be broadly categorized into specialized strategies (e.g., scooping \cite{cha2024high} \cite{he2021scooping} \cite{babin2019stable} \cite{babin2018picking} and prying \cite{zhang2022prying}) and environment-assisted approaches \cite{bimbo2019exploiting} \cite{hang2019pre} \cite{sarantopoulos2018human}.
Scooping typically requires specific fingertip designs \cite{cha2024high} \cite{he2021scooping} \cite{babin2019stable} \cite{babin2018picking}, whereas prying depends on the chamfers of objects \cite{zhang2022prying}.
Environment-assisted methods commonly exploit table edges through slide-to-edge strategies \cite{bimbo2019exploiting} \cite{hang2019pre} \cite{sarantopoulos2018human}.
In this work, for planar rigid objects (e.g., rulers), an environment-assisted approach was adopted, but it was extended beyond table edges by leveraging object edges (e.g., books) as additional supports, making more-flexible manipulation possible in cluttered, multiobject scenarios.

For deformable objects, prior studies often adapted strategies from cloth manipulation by creating graspable protrusions \cite{borras2020grasping}.
However, such methods do not account for the stiffness variations in paperlike materials.
In this study, how thickness influences graspability was analyzed explicitly.
For books, previous related works mainly focused on shelf organization \cite{yang2025planning} or reorientation by pushing \cite{huang2024toward}; instead, a prying-based strategy was used in this study for reliable grasping in tabletop settings.

\subsection{Exploiting Environmental Constraints}

Environmental constraints play a crucial role in robotic manipulation by reducing uncertainty and simplifying control \cite{eppner2015exploitation}.
Planar support from tabletops is widely used for rigid and deformable objects \cite{salvietti2015modeling} \cite{borras2020grasping}, whereas additional constraints, such as edges, walls, and grooves, further enhance manipulation capabilities \cite{sarantopoulos2018human} \cite{sundaram2020environment} \cite{liang2021learning} \cite{turco2021grasp} \cite{ding2024preafford}.
A representative example is edge-based grasping of flat objects, where objects are pushed to table boundaries to create graspable configurations \cite{bimbo2019exploiting} \cite{hang2019pre} \cite{sarantopoulos2018human}.
Extending this idea, in this study, it was demonstrated that object–object interactions can also be exploited as effective constraints.
In particular, book edges were utilized to assist the grasping of planar rigid objects, highlighting the broader potential of environmental resources in multiobject manipulation.

\begin{figure*}[htbp]
  \centering
  \includegraphics[width=15cm]{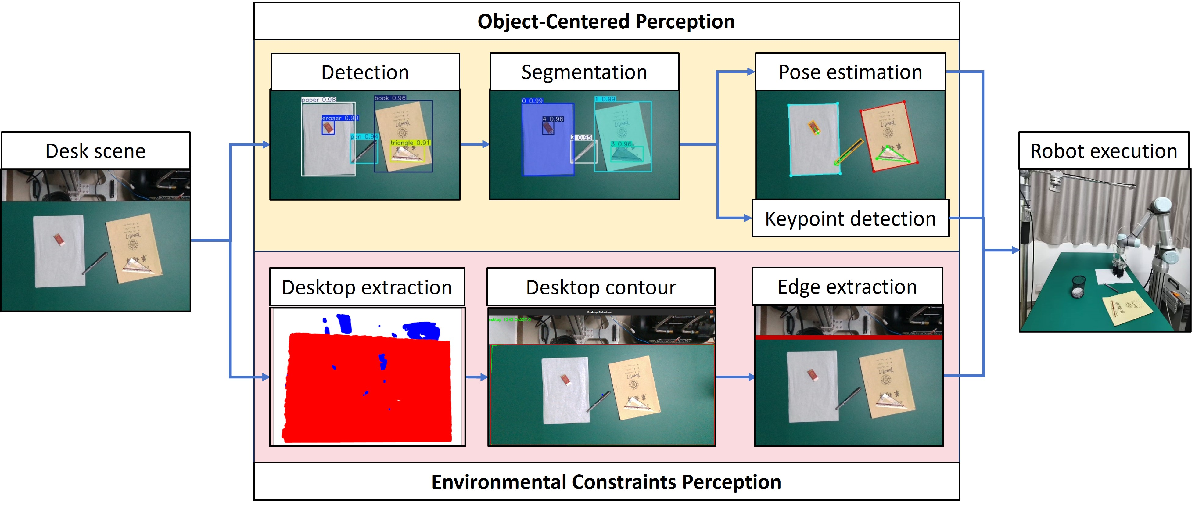}
  \caption{Perception pipeline for desktop organization. 
  The RGB-based branch first employs YOLO for object category detection, followed by SAM for object segmentation, and subsequently estimates object poses and keypoints for different objects according to task requirements.
  The point-cloud-based branch performs denoising, extracts the tabletop plane using RANSAC, computes its convex hull and projected contour, and finally obtains the table edges.}
  \label{visual_pipeline}
\end{figure*}

\section{PROPOSED METHODS}

This task involves objects of varying types, sizes, stiffnesses, and dimensionalities. 
To facilitate their perception and manipulation, the objects were classified into three primary categories: small objects, defined as objects that are small and can be grasped directly (e.g., a pen, an eraser, and a lead case), planar rigid objects (i.e., rulers, including both a straight ruler and a triangular ruler), and planar deformable objects (a sheet of paper and a book).

\subsection{Perception Pipeline} 

The visual requirements vary according to the three object categories. 
For small objects that can be grasped directly, acquiring the grasping pose is sufficient. 
For planar rigid objects (e.g., rulers) and planar deformable objects (e.g., paper and books), which are not amenable to direct grasping, it is necessary to extract their keypoints to infer the subsequent manipulation pose. 
Because environmental constraints can assist in grasping such planar objects, the vision system must also capture them.
Overall, the vision system must acquire all available cues in the scene, including object poses, keypoints, and environmental constraints.

The method to obtain these poses and keypoints is shown in Fig. \ref{visual_pipeline}. 
First, YOLO11 is used to identify the categories of different objects in the desktop scene.
Then, using the bounding boxes as prompts for the segmentation algorithm (SAM2.1), the outer contours of each object are obtained.
Next, for small objects, the minimum bounding rectangle of the contour provides both the position (center point) and the orientation (angle). 
For planar rigid objects (rulers) and planar deformable objects (paper and books), which have distinct shape features, a geometric model-based approach is adopted: convex hull detection combined with polygon approximation to extract polygon vertices as keypoints.
These keypoints are only two-dimensional points, and their three-dimensional coordinates must be obtained by combining depth information.

To train the visual recognition model (i.e., YOLO11), a dataset was also built specifically for objects in a desk scene.
Although datasets COCO 128, Objects 365, and the Stationery Dataset contain data about stationery, these datasets only include some common stationery items and do not cover all desk scene objects (e.g., triangular rulers and lead cases). 
Moreover, most of them are single-object scenes, lacking cluttered scenes with multiple objects, similar to those in real life.
To enhance task robustness, a dataset was constructed comprising both single-object and multiobject scenes. 
The dataset includes 2820 images for single-object scenes and 7397 images for multiobject scenes. 
During the acquisition of single-object images, the imaging angles, distances, and backgrounds were varied systematically, while external distractors were also introduced. 
For multiobject scenes, the dataset further covers common object combinations and varying degrees of overlap between objects. 

For planar rigid objects (e.g., rulers), environmental constraints, such as the table edge, are necessary for successful grasping; therefore, an environmental constraint detection algorithm was developed (as shown in Fig. \ref{visual_pipeline}).
A depth camera first captures synchronized RGB and point cloud data of the scene.
After downsampling and noise filtering, RANSAC is applied to segment the table plane from the point cloud. 
The 3D convex hull of this plane is projected onto the 2D image to generate a polygonal workspace contour.
Based on this contour, a clear desktop edge is fitted, which is then used to select the optimal grasping position for planar rigid objects. 
This depth-based approach ensures robustness against variations in surface color and texture.

\subsection{Multiprimitives for Diverse Objects}

\subsubsection{Contact Grasping}
\begin{figure}[htbp]
  \centering
  \includegraphics[width=8.5cm]{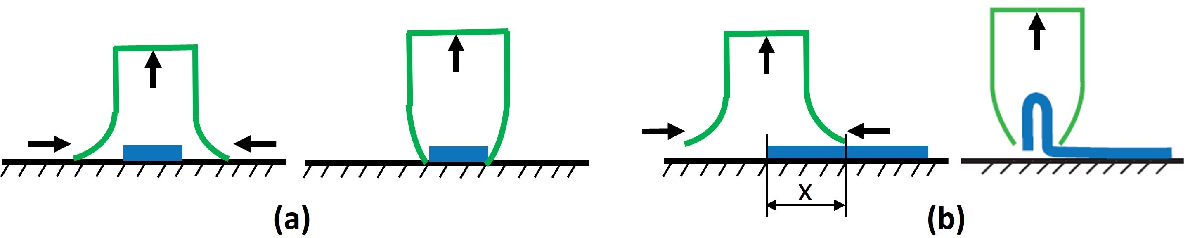}
  \caption{Contact-based grasping primitives.
  Illustration of grasping (a) small objects and (b) deformable objects.}
  \label{paper_grasping}
\end{figure}

The contact-grasping method was proposed in previous work \cite{eppner2015exploitation}. 
Its workflow is as follows: first, the grippers open, aligning with the grasping pose of the object. The fingertips initially contact the tabletop; then, while maintaining contact with the tabletop, they close until the fingers touch the object and grasp it (Fig. \ref{paper_grasping} (a)). 
It has been shown to have a higher success rate in grasping small-diameter or short objects, which aligns well with the requirements for grasping small objects in the task in this study.
In experiments, the success rate of this strategy in grasping different small items, as well as its robustness to grasping position errors, was tested.
In addition, this method was applied to the grasping of flexible paper. 
Unlike grasping rigid objects, grasping flexible paper involves deformation (Fig. \ref{paper_grasping} (b)), which is closely related to the grasping position x and paper thickness (characterized by GSM specifications). 
Therefore, in Section IV, the workspace of this grasping strategy for paper is presented, defined by the feasible ranges of these parameters.

The key to this strategy is maintaining contact between the fingertips and the tabletop during the fingertip closing process. 
This can be achieved using compliant control methods, passive joints, or soft grippers. 
Unlike in previous work, soft grippers were chosen to avoid the complexity of control algorithms and mechanisms while protecting paper materials from damage.
Although they offer lower stiffness, it is sufficient for everyday tasks (especially for the lightweight stationery objects in this task).

\subsubsection{Push Grasping}

\begin{figure}[htbp]
  \centering
  \includegraphics[width=8.5cm]{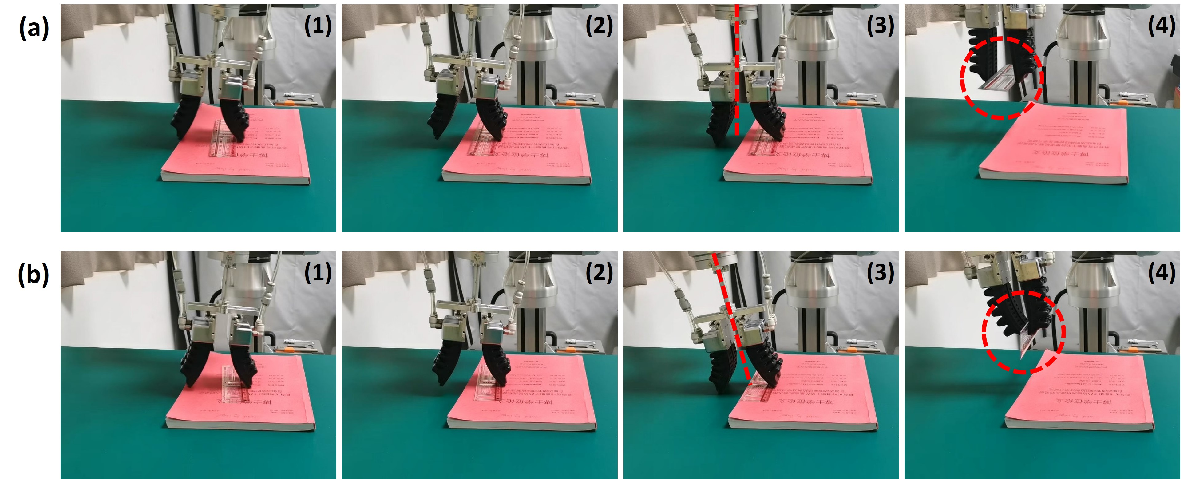}
  \caption{Push-grasping primitive for retrieving a ruler from a book.
  (a)–(b) Examples under different initial poses of the ruler.
  (1)–(4) correspond to the pushing start pose $P_{push}^s$, pushing end pose $P_{push}^e$, grasping pose $P_{grasp}$, and the lifted state of the object under different grasping poses.}
  \label{ruler_grasping}
\end{figure}

According to \cite{bimbo2019exploiting}, two types of edge-based grasping method exist: (1) continuous slide and grasp and (2) pivot and regrasp.
Because the ruler is a very thin object, continuous slide and grasp is more suitable for grasping, and this strategy is also more time efficient.
Unlike in previous studies, in this multiobject task, the ruler can utilize not only the edge of the tabletop but also the edges of other objects, such as the edge of a book.
As shown in Fig. \ref{ruler_grasping}, the grasping process is as follows: first, the pushing initial pose $P_{push}^s$ is determined; then, the object is pushed to the edge of the supporting surface, reaching pose $P_{push}^e$. Finally, the grasp $P_{grasp}$ is performed.
The grasp pose, whether without an angle or with an angle, determines whether the ruler is grasped by the two edges (Fig. \ref{ruler_grasping} (a)(4)) or the two sides (Fig. \ref{ruler_grasping} (b)(4)); the latter is more stable.

\begin{algorithm}
\caption{Push–grasp pose generation}
\label{alg:push-grasp}
\SetKwFunction{IsOnBook}{IsOnBook}\SetKwFunction{PushDir}{PushDirection}
\SetKwFunction{FootPerp}{Foot}\SetKwFunction{EdgeCent}{EdgeCenters}
\SetKwFunction{MinRect}{MinRect}\SetKwFunction{CrtPose}{CreatePose}
\SetKwFunction{RotDir}{RotationDirection}\SetKwFunction{Rot}{Rotate}

\KwIn{object $obj$, book $B$, desktop $D$}
\KwResult{push poses $P_{push}^{s}, P_{push}^{e}$, grasp pose $P_{grasp}$}

\tcp{1. Obtain surface and direction}
$S \leftarrow \begin{cases} 
    B & \text{if } \text{IsOnBook}(obj.pos, B.corners) \\
    D & \text{otherwise}
\end{cases}$\;
$d \leftarrow$ \PushDir{$obj.\theta$, $S.\theta$}\;

\tcp{2. Compute grasp position}
$p_g^{2d} \leftarrow$ \FootPerp{$S.corners$, $d$, $obj.pos$}\;

\tcp{3. Compute push position}
$E \leftarrow$ \EdgeCent{\MinRect{$obj.corners$}}\;  
$p_p^{2d} \leftarrow \arg\max_{p^{2d} \in E} \|p^{2d} - p_g^{2d} \|$\;

\tcp{4. Generate poses}
$P_p \leftarrow$ \CrtPose{$p_p^{2d}$, $obj.z$, $obj.\theta$}\;
$P_g \leftarrow$ \CrtPose{$p_g^{2d}$, $obj.z$, $S.\theta$}\;

\tcp{5. Adjust gripper orientation}
$r \leftarrow$ \RotDir{$obj.\theta$, $p_p^{2d}$, $p_g^{2d}$}\;
$P_{push}^{s} \leftarrow$ \Rot{$P_p$, $\theta_p$, $r$}\;
$P_{push}^{e} \leftarrow$ \CrtPose{$P_{push}^{s}.ori$, $P_g.pos$}\;
$P_{grasp} \leftarrow$ \Rot{$P_g$, $\theta_g$, $r$}\;

\Return $P_{push}^{s}$, $P_{push}^{e}$, $P_{grasp}$
\end{algorithm}

The execution of the push–grasp primitive requires determining the pushing initial pose $P_{push}^s$ and final pose $P_{push}^e$ and the grasp pose $P_{grasp}$. The calculation process is detailed in Algorithm 1.
First, visual information about the ruler (including straight and triangular rulers) and potential supporting surfaces (books or the desktop) is acquired, based on which the supporting surface $S$ (book or desktop) and push direction $d$ (along the long or short side) are determined.
Given the supporting surface and push direction, the foot of the perpendicular from the position of the ruler to the nearest side of the supporting surface is calculated as the grasp position $p_g^{2d}$.
The push position $p_p^{2d}$ is the center point of the long side of the ruler that is closest to the grasp position $p_g^{2d}$. 
Then, based on the push position, grasp position, object orientation, and supporting surface orientation, two reference poses—$P_p$ (push reference pose) and $P_g$ (grasp reference pose)—are defined.
Adding an angle to these two reference poses yields the push pose $P_{push}^s$ and grasp pose $P_{grasp}$. 
The angle of the push pose $\theta_p$ facilitates the push action, whereas the angle of the grasp pose $\theta_g$ is used to grasp the two sides of the ruler (Fig. \ref{ruler_grasping} (b)(4)). 
$P_{push}^e$ (Fig. \ref{ruler_grasping} (b)(2)) is the final pose of the push action and the transition between $P_{push}^s$ and $P_{grasp}$, depending on the orientation of $P_{push}^s$ and the position of $P_g$.

\subsubsection{Pry Grasping}

\begin{figure}[htbp]
  \centering
  \includegraphics[width=8.5cm]{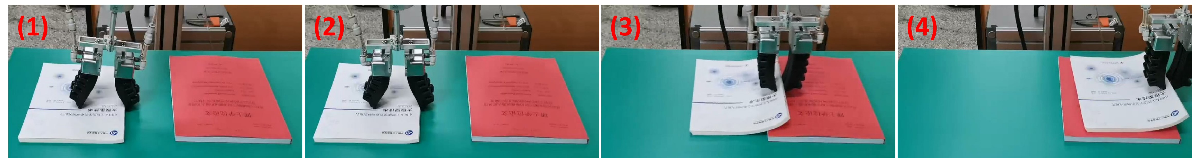}
  \caption{Pry-grasping primitive.
  (1) Contact phase; (2) Prying phase; (3) Pulling phase; (4) Book stacking.}
  \label{pry_grasping}
\end{figure}

For deformable objects, such as books, a grasping method similar to one developed previously \cite{zhang2022prying} can be used: pry grasping. 
The workflow is shown in Fig. \ref{pry_grasping}. 
First, the gripper, in its open state, contacts the book, with the left fingertip touching the cover and the right fingertip touching the bottom edge of the spine. The angle of the gripper is determined by the thickness of the book. After contact, the grippers close (Fig. \ref{pry_grasping} (a)(1)). 
Next, the prying phase begins, with the grippers rotating at a certain angle $\alpha$, which is defined as the prying angle and is also the angle at which the spine is pried up (Fig. \ref{pry_grasping} (a)(2)).
Subsequently, the pulling phase begins, with the grippers moving to the right. 
During the pulling phase, the left fingers slide along the book, while the book simultaneously slides along the inside of the right fingers under the applied force, until both fingers stably grasp the spine of the book (Fig. \ref{pry_grasping} (a)(3)). In other words, a stable grasp is achieved during the pulling phase. 
Finally, the book is placed on top of another book and stacked neatly (Fig. \ref{pry_grasping} (a)(4)).

This study differs from a previous one \cite{zhang2022prying} in two main aspects. 
First, soft grippers were employed, and the passive compliance of the fingers was leveraged, rather than relying on multijoint fingers and complex force control algorithms. 
Second, the target object in the task was a large-area deformable object—a book—and the initial tilt angle of its spine (Fig. \ref{primitive3_analysis} (a)) is more pronounced than the chamfer of the object, making it easier to pry.
The most important factors for the successful execution of the pry-grasping primitive for books include the physical properties of the book, the grasping force, and the prying angle $\alpha$. 
With the operating air pressure of the gripper fixed, the influences of book thickness and prying angle on the success rate of pry grasping were investigated, as detailed in Section \ref{sec:experiments}.

\subsection{Desk Organization Task Planner}

Based on the initial state of the desktop scene, a dedicated task planner was developed.
The planner is formulated using object-centered primitives, which are built upon the motion primitives defined in the previous section.
Specifically, for small objects and planar rigid objects, the previously defined contact grasp and push–grasp primitives are adopted, respectively. 
In addition, the target placement pose of each object relative to the pen holder is explicitly encoded using a placement primitive.
For planar deformable objects, paper is grasped using the contact grasp primitive, and books are handled using the pry primitive. 
Furthermore, when paper or books are involved, an additional reorientation primitive is required to align their edges parallel to the desk boundaries for neat placement.
After the appropriate object-centered primitives are determined, the desktop is organized according to the spatial relationships among objects.
The process begins with small objects, including pens, erasers, and lead cases, which are typically placed on the top layer after use.
Subsequently, the planner determines the execution order between planar rigid objects and deformable objects based on the spatial dependency between the ruler and the books. 
If the ruler is located on a book, it can be grasped directly and placed by leveraging the book edge. Otherwise, deformable objects are organized first to reduce potential obstacles for the push–grasp operation of the ruler.
Finally, before executing object-centered primitives, the planner checks for adjacent objects. 
If interference is detected, a push action is performed to separate the objects, ensuring reliable execution of the subsequent manipulation.

\section{EXPERIMENTS AND RESULTS}
\label{sec:experiments}

\subsection{Experimental Setup and Materials}

\begin{figure}[htbp]
  \centering
  \includegraphics[width=8.5cm]{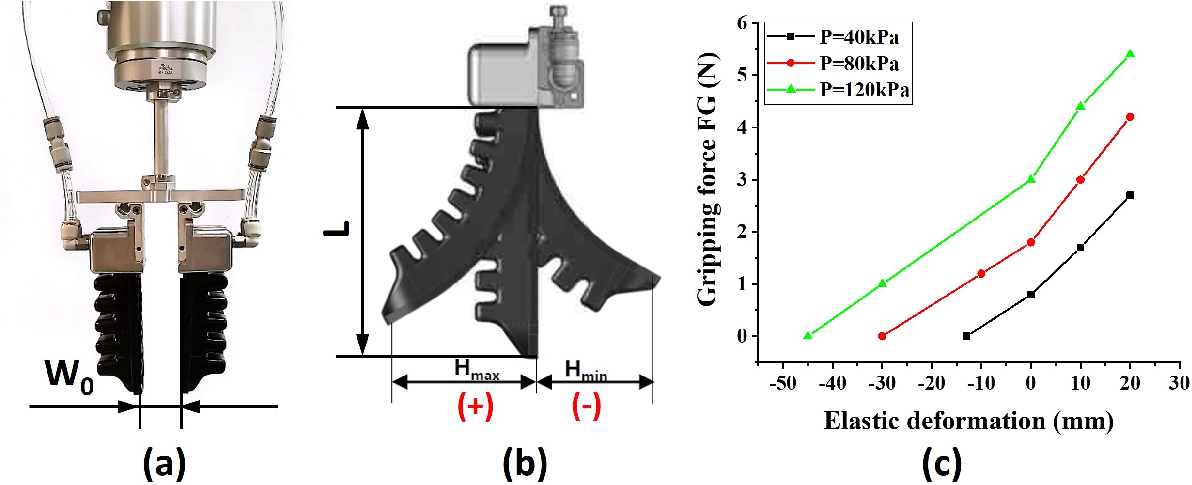}
  \caption{Workspace and gripping force of the finger.
  (a) Gripper mounted on the end of the robot.
  (b) The specifications of the finger: L = 83 mm, Hmax = 33 mm (-80 kPa), Hmin = 39 mm (100 kPa). 
  (c) The diagram shows the relationship between gripping force FG and the deformation of the finger when grasping objects.}
  \label{Gripper}
\end{figure}

As shown in Fig. \ref{visual_pipeline}, the entire system is a robotic system based on a soft gripper. In the experiment, a four-finger gripper was used (Rochu GC-4FMA6V5/LS1).
As shown in Fig. \ref{Gripper} (a), the finger spacing $W_0$ is 30 mm. The finger parameters are finger length $L$ of 83 mm, maximum open deformation $H_{max}$ of 33 mm (-80 kPa), and maximum closed deformation $H_{min}$ of 39 mm (100 kPa) (Fig. \ref{Gripper} (b)).
The relationship between the deformation of a single finger and the resulting horizontal force under pressure is shown in Fig. \ref{Gripper} (c).
Furthermore, the robot used was an UR5e, and the camera was an Intel RealSense D415, positioned approximately 0.74 m above the table.
The desktop computer used in the experiment had an Intel i7 2.9 GHz CPU and an NVIDIA GeForce RTX 3080 GPU.
The experimental algorithm was implemented within the ROS framework, and MoveIt was used for motion planning of the robotic manipulator.

\begin{figure}[htbp]
  \centering
  \includegraphics[width=8.5cm]{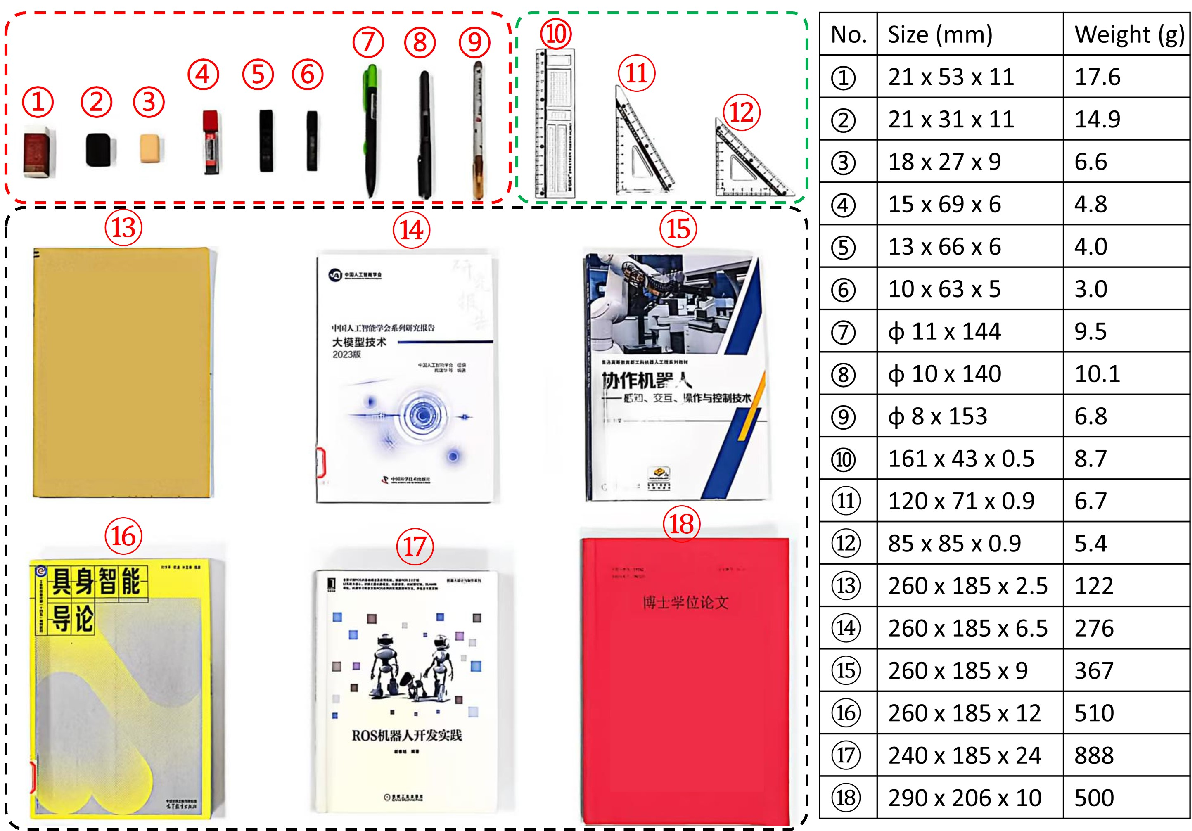}
  \caption{Objects used in primitive experiments.
  Red dashed boxes indicate tested small objects; green dashed boxes indicate tested planar rigid objects (rulers); black dashed boxes indicate tested planar deformable objects (books).
  The table on the right lists the dimensions and masses of the corresponding items. }
  \label{Materials}
\end{figure}

The test items used in the organization experiments for different object categories—namely small objects, planar rigid objects, and planar deformable objects—are shown in Fig. \ref{Materials}. 
The table on the right lists the dimensions and masses of the corresponding items. 
For planar deformable objects, paper items are not shown in the figure, as they exhibit no visual differences; their primary distinction lies in thickness, which is mainly determined by the paper specification measured in grams per square meter (GSM). 
The items listed in the figure were used solely for testing the primitives. 
In the dataset construction and the final desktop organization task, the number of object instances far exceeded those shown in the figure.


\subsection{Small Object Manipulation}

\begin{figure}[htbp]
  \centering
  \includegraphics[width=8.5cm]{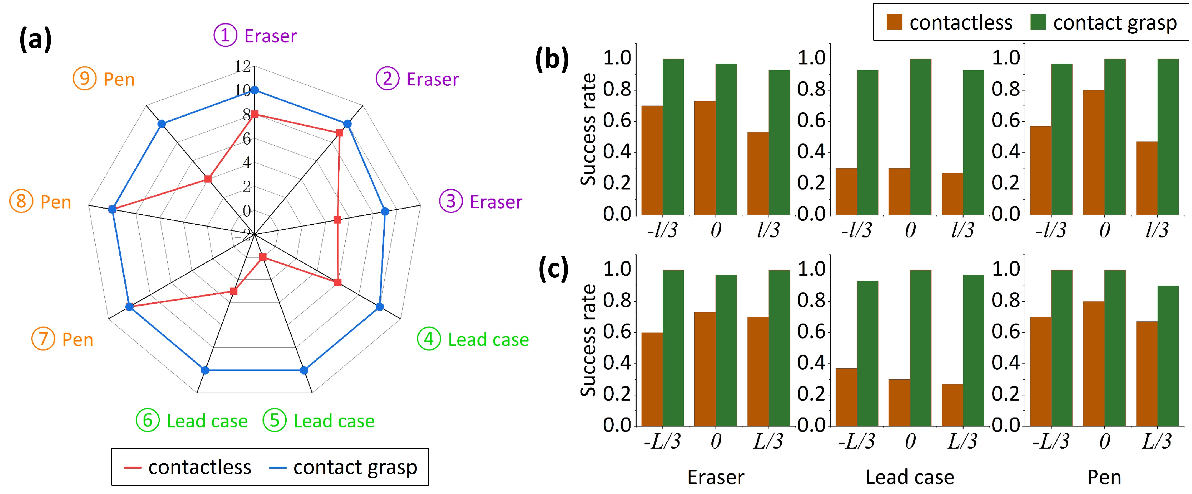}
  \caption{Success rate of grasp-and-place for small objects.
  (a) Success rate distribution of different object types and sizes under contact and non-contact grasping;
  (b) Comparison of success rates under lateral offsets;
  (c) Comparison of success rates under longitudinal offsets.}
  \label{primitive1_results}
\end{figure}

In the primitive testing for object-centered tasks, the “success rate” refers to the completion rate of object organization, which for small objects specifically corresponds to the success rate of final placement.
First, the impacts of contact grasping and noncontact grasping on small objects of different categories and sizes were evaluated.
Three representative objects of varying sizes were selected from each of the three small-object categories, as shown in Fig. \ref{Materials}.
For each object instance, the success rate was tested for both contact grasping (grasping depth: 1.5 $\times$ height) and noncontact grasping (grasping depth: 0.5 $\times$ height). Each object was tested ten times, and the results are shown in Fig. \ref{primitive1_results} (a).
The results show that, regardless of the object category and size, the success rate of contact grasping was consistently close to $100\%$. Only Eraser 3 failed once during placement because of jamming in the gripper.
For noncontact grasping, the performance is highly sensitive to object height. For objects with heights of 9 mm or less, the success rate was $\leq$ 6/10.
There are several reasons for the failure of noncontact grasping, including the limited depth accuracy of the camera, object chamfers, and fingertip geometry.

Subsequently, lateral (along the minor dimension $l$) and longitudinal (along the major dimension $L$) offset variables were introduced to evaluate the robustness of contact and noncontact grasping against positional deviations.
The offset magnitude was set to one-third of the corresponding object dimension in each direction. 
For each condition, ten trials were conducted per object, and the results were aggregated within each object category for comparison, as shown in Fig. \ref{primitive1_results} (b) and (c).
The results further confirm the previous finding that contact grasping exhibits strong robustness to object height. 
In addition, contact grasping demonstrates superior robustness to positional offsets, achieving a minimum success rate of $90\%$ (27/30).
Failures in the offset cases are mainly caused by the additional torque induced by the positional deviation, which alters the object pose after grasping and may lead to placement failure.
Finally, for different object types, the success rates under positive and negative offsets exhibited a certain degree of symmetry in the lateral and longitudinal directions, which is primarily attributed to the axial symmetry of the object geometry.

\subsection{Planar Rigid Objects Manipulation}

\begin{figure}[htbp]
  \centering
  \includegraphics[width=8.5cm]{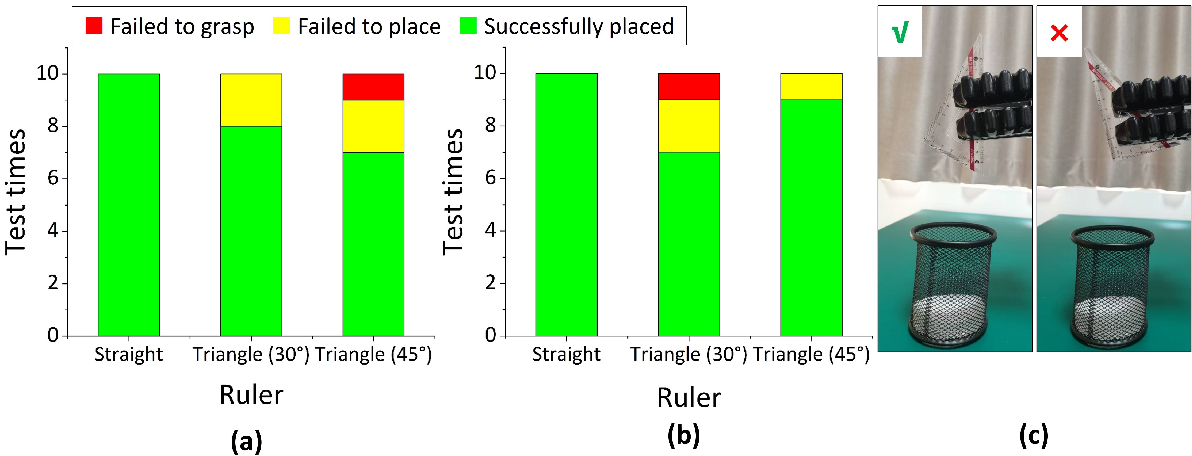}
  \caption{Success rate of grasp-and-place for rulers on different supporting surfaces.
  (a) Rulers on books;
  (b) Rulers on the tabletop;
  (c) Examples of successful and failed placements of triangular rulers.}
  \label{primitive2_results}
\end{figure}

The success rates of grasping and placing different types of rulers were evaluated in two scenarios: on a book and directly on the table.
The tested ruler types included straight rulers, 30-degree triangular rulers, and 45-degree triangle rulers, with random initial poses for all rulers.
Ten trials were conducted under each condition, and the resulting success rates are summarized in Fig. \ref{primitive2_results}.
The straight ruler achieved a $100\%$ success rate in both scenarios.
In contrast, failures occurred for the two triangular rulers, including grasping failures (red) and placement failures (yellow).
The ratios of placement failures to grasping failures were 4:1 on the book and 3:1 on the table, indicating that placement was the dominant source of failure.
The relatively few grasping failures are mainly attributed to depth estimation errors and inaccuracies in table edge detection.
Placement failures primarily arise from the nonrectangular geometry of triangular rulers, which leads to pose uncertainty after grasping.
As a result, the ruler edges or corners may extend beyond the pen holder boundary during placement, causing failure (Fig. \ref{primitive2_results} (c)).

\subsection{Planar Deformable Object Manipulation}

\begin{figure}[htbp]
  \centering
  \includegraphics[width=8.5cm]{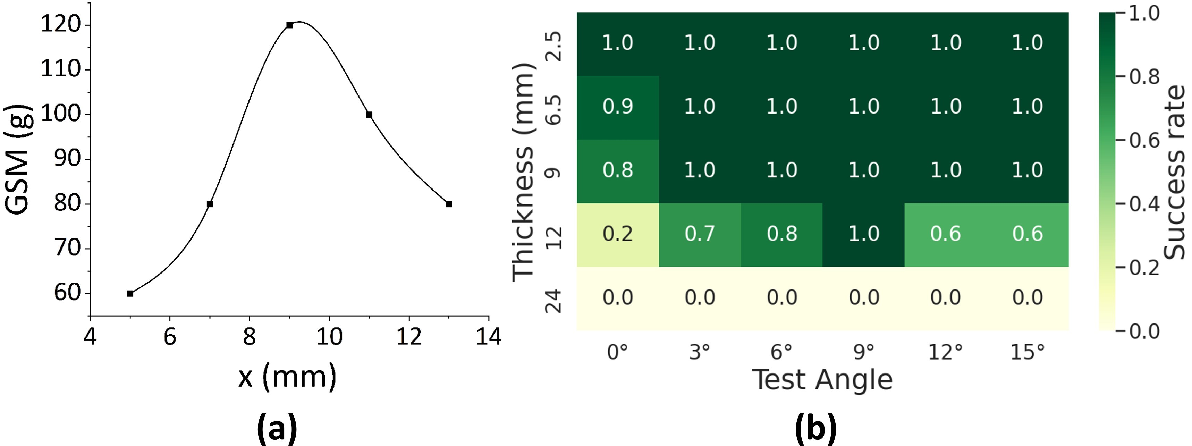}
  \caption{Organization experiments with planar deformable objects.
  (a) Critical specifications of paper for different grasping locations;
  (b) Distribution of grasp-and-stack success rates for books under varying prying angles and thicknesses.}
  \label{primitive3_results}
\end{figure}

For 2D deformable thin paper, contact grasping is adopted, where the required deformation force is influenced by the deformation length and paper thickness.
At a given grasping position, the deformation force increases with paper thickness (i.e., GSM), whereas the grasping force remains constant, thereby leading to a critical graspable thickness.
Different grasping positions and their corresponding critical paper specifications, defined as the maximum thickness achieving a $100\%$ success rate (10/10), were evaluated. The results are shown in Fig. \ref{primitive3_results} (a).
An edge effect was observed within the grasping range. When the grasping position was 90 mm (close to the gripper opening limit), the critical thickness reached its maximum (GSM = 120 g).
This can be attributed to two factors: (1) the grasping length reaches its maximum, i.e., the gripper opening limit, and (2) an edge-contact effect, where the fingers establish initial static contact with the paper edge prior to grasping.
Considering that common paper typically ranges from 60 to 80 g, a grasping position of 70 mm was selected, which was sufficient for reliable paper manipulation in the target task.

\begin{figure}[htbp]
  \centering
  \includegraphics[width=8.5cm]{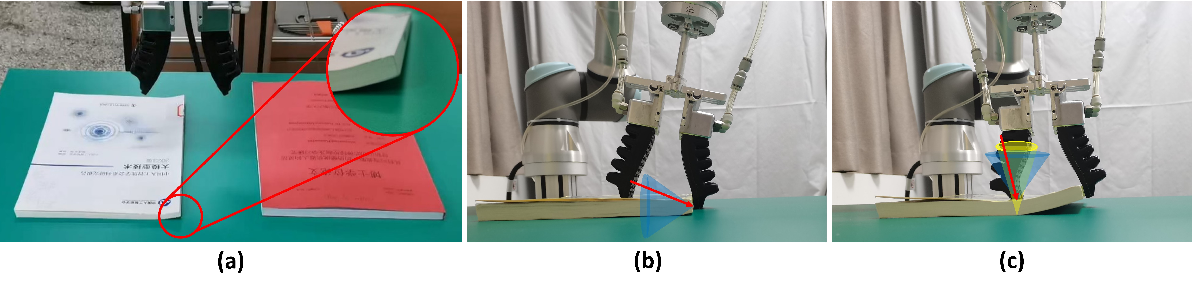}
  \caption{Book initial state and pry-grasping failure analysis. 
  (a) Initial spine tilt forms a natural gap with the supporting surface;
  (b) At 0° prying, the right-finger contact force remains within the friction cone, preventing the book from sliding along the finger; 
  (c) At 15° prying, the left-finger contact force also lies within the friction cone, causing initial pulling jamming.}
  \label{primitive3_analysis}
\end{figure}

For deformable books, pry primitives make prying–pulling–placement operations possible. The success rates of varying pry angles were evaluated for different thicknesses (10 trials per condition), as shown in Fig. \ref{primitive3_results} (b). 
The spine angle of these old books created a gap of 5 ± 2 mm between the spine and the table/surface (Fig. \ref{primitive3_analysis} (a)).
For 2.5 and 6.5 mm books, nearly $100\%$ success was achieved over $0^\circ$--$15^\circ$, and even at $0^\circ$ because of the inherent spine angle and low stiffness. 
At 9 mm, the success rate at $0^\circ$ dropped to $8/10$ because insufficient book sliding along the right finger led to a weak grasp (within the friction cone) (Fig. \ref{primitive3_analysis} (b)).
For 12 mm, the success rate was $\leq 8/10$ when $\theta < 6^\circ$ or $\theta > 12^\circ$ because large angles altered the left-finger force direction and hindered sliding (Fig. \ref{primitive3_analysis} (c)). The optimal angle was $9^\circ$ with $100\%$ success.
At 24 mm, the task failed for all angles owing to the increased weight and insufficient pulling force, which can be mitigated by a higher gripper pressure or a push–grasp strategy.

\subsection{Whole Framework Evaluation}

\begin{figure}[htbp]
  \centering
  \includegraphics[width=8cm]{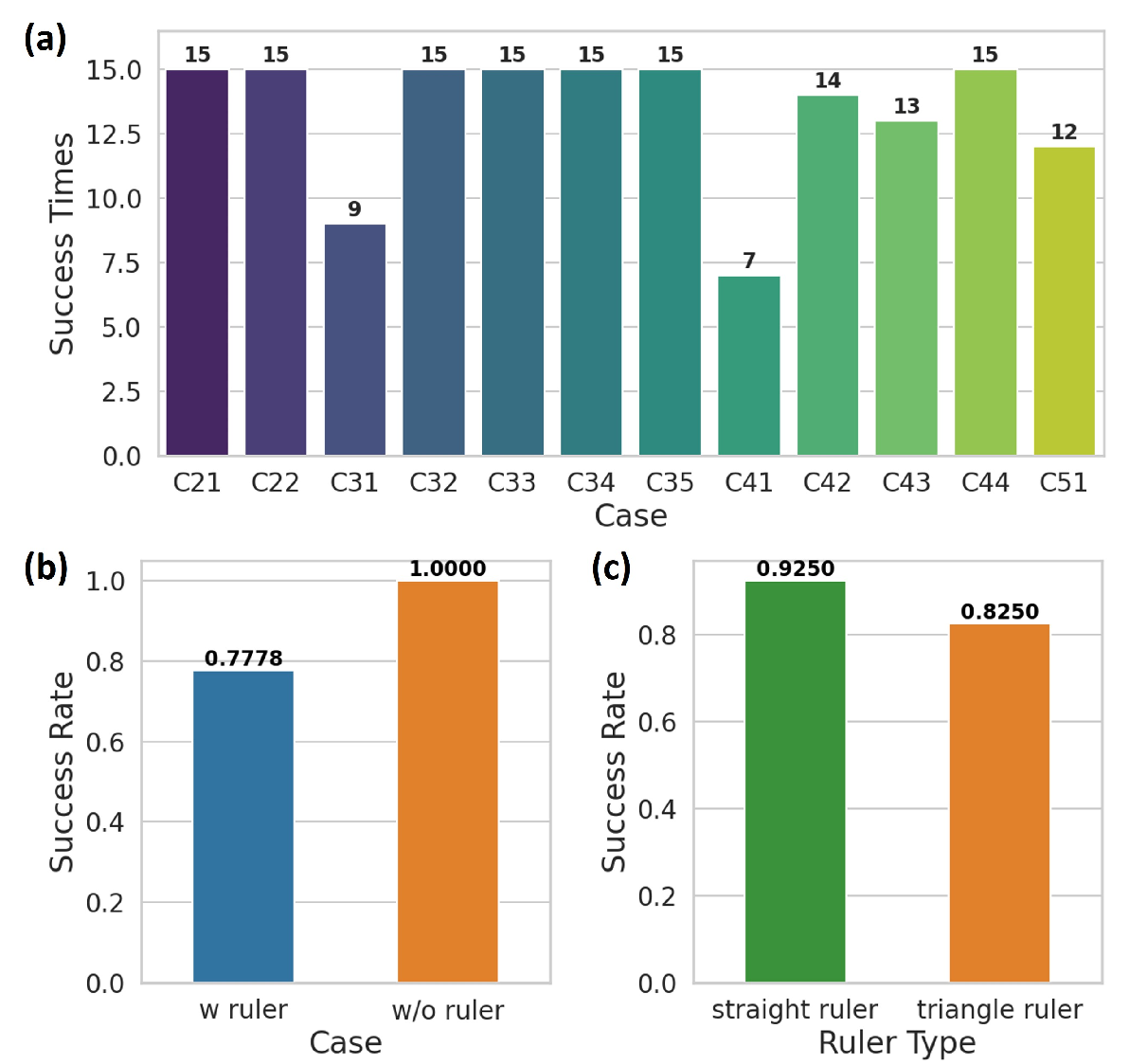}
  \caption{Overall task success rate and analysis for desktop organization.
  (a) Distribution of successful task completions across different scenarios;
  (b) Comparison between scenarios with and without rulers;
  (c) Comparison between scenarios with straight rulers and triangular rulers (excluding special cases of C311 and C411 rulers on paper).}
  \label{task_results}
\end{figure}

The overall task success rate was evaluated across 12 common object combinations involving multiple categories, comprising two to five object categories.
Each scenario is denoted as \texttt{Cxyz}, where $x$ is the number of categories, $y$ indexes the combination, and $z$ denotes one of three layout instances. 
All 36 scenarios are provided on the project website, and each was tested five times. 
When small objects were placed on nontable surfaces, the grasping depth was set to one object height to avoid cograsping the supporting object.
The results are shown in Fig. \ref{task_results} (a).
Overall, the success rate decreased with the number of object categories owing to the increased number of manipulation steps and stronger interobject interference. 
Two outliers, C31 and C41, achieved only 9/15 and 7/15, respectively. 
This was mainly caused by specific cases (C311 and C411, subcases of C31 and C41 respectively) in which a ruler was placed on paper; in such cases, the push–grasp primitive tended to cograsp the paper, leading to task failure.

Excluding the two special cases mentioned previously, failures were mainly associated with pens (three cases) and rulers (seven cases).
For pens, failures were primarily caused by inaccurate grasping depth: reflective surfaces degrade depth estimation, whereas curved geometry makes shallow grasps unstable and deeper grasps prone to cograsping the underlying paper.
Because rulers dominate the failure cases, their impact was analyzed further. As shown in Fig. \ref{task_results} (b), scenarios without rulers achieved $100\%$ success, whereas those with rulers dropped to $77.78\%$. 
All pen-related failures occurred in the ruler-involving scenarios, indicating coupled interference effects.
To isolate the effect of ruler type, C311 and C411 were excluded, and scenes with straight rulers and triangular rulers were compared. The success rates were $92.5\%$ and $82.5\%$, respectively (Fig. \ref{task_results} (c)). 
Grasping failures are mainly caused by depth estimation and table edge detection errors, while triangular rulers also cause shape-induced placement failures.
Overall, these results indicate that the physical properties of an object, including the material reflectivity and geometric structure, are crucial factors influencing task success.

\section{DISCUSSIONN \& CONCLUSION}
A desk organization task involving heterogeneous objects was investigated, and task-oriented manipulation primitives were developed for different categories.

\textbf{Small object manipulation:} 
For small objects, contact-based grasping was adopted, and its robustness with respect to the object height and grasping location was demonstrated. 
Although similar primitives have been explored in prior work \cite{eppner2015exploitation}, in this study, their effectiveness was validated in a realistic desk scenario, and the key factors affecting success were analyzed systematically. 
In practice, object–object interactions (e.g., small items placed on paper) introduce additional challenges, where perception errors can lead to unintended cograsping. 
This can be mitigated either by improving the depth-sensing accuracy or by adjusting the task strategy, such as temporarily relocating supporting objects to expose targets.

\textbf{Planar rigid-object manipulation:}
For planar rigid objects, such as rulers, environment-assisted manipulation was leveraged not only by exploiting the desk edge \cite{bimbo2019exploiting} \cite{hang2019pre} \cite{sarantopoulos2018human} but also by extending to the edges of other supporting objects.
This highlights that object–object contact can serve as a useful constraint in multiobject settings. 
However, although soft grippers provide adaptability, the uncertainty in in-hand object pose introduces challenges for precise placement, especially for irregular shapes, such as triangular rulers. 
Future improvements may incorporate in-hand pose estimation or reorientation strategies to enhance placement accuracy.

\textbf{Planar deformable object manipulation:}
For planar deformable objects, including paper sheets and books, different strategies were designed based on the physical properties of the objects. 
For paper, the relationships between thickness, grasping location, and success rate were analyzed to provide practical guidelines for stable grasping. 
For books, a prying-based primitive using a soft gripper was employed, which avoids complex force control by exploiting passive compliance and object deformability. 
Furthermore, for stacking tasks, full lifting is not necessary; pulling with table support can reduce grasping force requirements. 
As object stiffness increases, however, rigid-object strategies, such as leveraging desk edges, become more suitable.

\textbf{Desktop organization:}
Building upon these primitives, a task planner was developed for real-world desk organization involving structured operations, such as sorting and stacking. 
Compared with conventional tabletop rearrangement tasks \cite{hu2025planning} \cite{gao2022fast}, the considered scenario requires coordinated manipulation of diverse and hard-to-grasp objects. 
Experimental results show that task success decreases as the number of objects increases because of longer action sequences and more-complex spatial relationships (e.g., adjacency and stacking). 
Performance can be further improved through both hardware (e.g., higher-precision sensing and tactile feedback) and algorithmic enhancements (e.g., failure recovery and robust perception for reflective objects).

Overall, this work demonstrates that environmental constraints—including table surfaces, edges, and object–object contacts—can be exploited systematically across different manipulation stages to improve robustness and efficiency. 
The proposed framework provides a practical reference for desk organization tasks involving rigid and deformable objects and highlights the broader potential of leveraging interobject constraints in multiobject manipulation.













\bibliographystyle{IEEEtran} 
\bibliography{IEEEabrv, IEEEexample}

\end{document}